\pdfoutput=1

\documentclass[10pt,twocolumn,letterpaper]{article}

\PassOptionsToPackage{table}{xcolor}

\usepackage{paperstyle}

\definecolor{linkblue}{rgb}{0.21,0.49,0.74}
\usepackage[pagebackref,breaklinks,colorlinks,allcolors=linkblue]{hyperref}

\title{OT-MeanFlow3D: Bridging Optimal Transport and Meanflow for Efficient 3D Point Cloud Generation}

\author{
Elaheh Akbari\textsuperscript{1} \quad
Shansita Sharma\textsuperscript{2} \quad
Ping He\textsuperscript{1} \quad
Ahmadreza Moradipari\textsuperscript{3} \\
Kyungtae Han\textsuperscript{3} \quad
Hamed Pirsiavash\textsuperscript{4} \quad
Yikun Bai\textsuperscript{5,}\footnotemark[1] \quad
Soheil Kolouri\textsuperscript{1,}\thanks{Corresponding authors.} \\[2pt]
{\small \textsuperscript{1}College of Connected Computing, Vanderbilt University \quad
\textsuperscript{2}Department of Computer Science, University of Southern California} \\
{\small \textsuperscript{3}Toyota Motor North America, InfoTech Labs \quad
\textsuperscript{4}Department of Computer Science, University of California, Davis} \\
{\small \textsuperscript{5}Department of Mathematics, Purdue University} \\
{\tt\small elaheh.akbari@vanderbilt.edu} \\[-8pt]
}

\usepackage{orcidlink}

\usepackage{booktabs}
\usepackage{multirow}
\usepackage{threeparttable}

\usepackage{xcolor}
\usepackage{algorithm}
\usepackage[noend]{algpseudocode}

\algrenewcommand\algorithmiccomment[1]{%
  \hfill{\footnotesize\textcolor{blue}{\textit{//~#1}}}%
}

\usepackage{amsmath}

\usepackage{pifont}

\makeatletter
\g@addto@macro\normalsize{%
  \setlength\abovedisplayskip{6pt plus 2pt minus 2pt}%
  \setlength\belowdisplayskip{6pt plus 2pt minus 2pt}%
  \setlength\abovedisplayshortskip{0pt plus 2pt}%
  \setlength\belowdisplayshortskip{4pt plus 2pt minus 2pt}}
\makeatother

\begin{document}
\maketitle
\begin{abstract}
Flow-matching models have recently emerged as a powerful framework for generative modeling, including 3D point cloud synthesis. However, their deployment is limited by the need for multiple sequential sampling steps at inference time. MeanFlow enables single-step generation by learning an average velocity field, but its quality depends heavily on the geometry of the training pairs. In this work, we propose an Optimal Transport-enhanced MeanFlow framework (\textbf{OT-MF3D}) for efficient and accurate 3D point cloud generation and completion. Beyond incorporating OT into MeanFlow, our method introduces a latent-conditioned training algorithm that decouples shape-level sampling from point-level matching. This enables exact point-level OT within each conditioned shape while performing shape-level transport in a compact latent space, avoiding the expensive nested OT computation in point-cloud space. Experiments on standard benchmarks show improved generation and completion quality over fast baselines, with substantially reduced training and inference costs compared with diffusion and other flow-based models.
\end{abstract}

\section{Introduction}
\label{sec:intro}

\begin{figure}
  \centering  
  \includegraphics[width=\columnwidth]{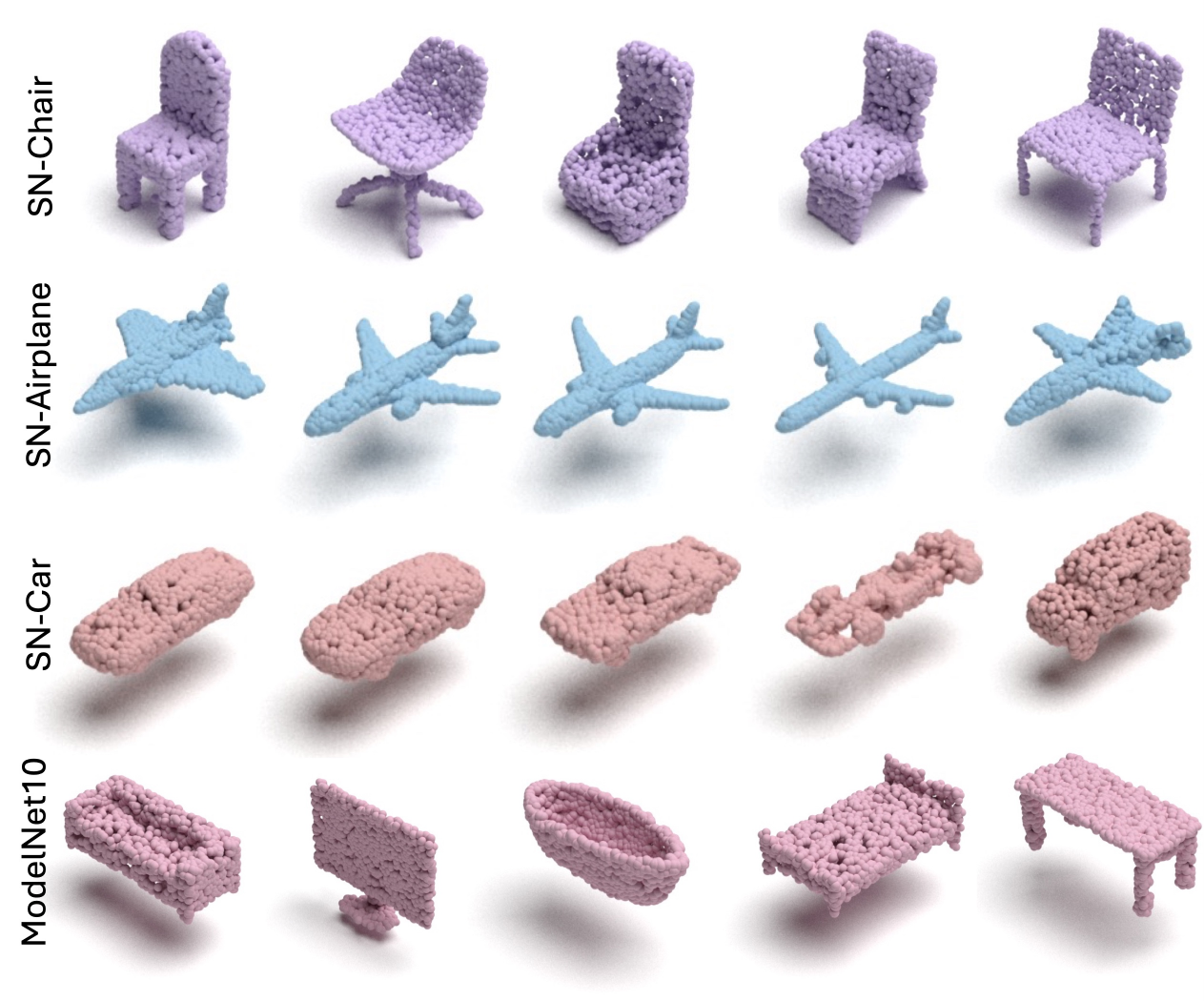}
  \caption{One-step generated samples for SN and ModelNet-10 generated within 0.04 seconds.}
  \label{fig:gen_samples}  
\end{figure}

Generating realistic 3D point clouds has broad applications in shape synthesis, 3D reconstruction, digital content creation, robotics, and autonomous systems \cite{guo2020deep}. However, modeling their distributions is challenging because point clouds provide unordered, irregular, and often sparse representations of complex geometry. Reliable generative models can address these limitations through tasks such as shape completion and point-cloud densification, with important applications including the enhancement of sparse LiDAR scans for 3D object detection. Unlike voxel- or mesh-based approaches, point-based models generate coordinates directly, avoiding intermediate representations and their associated computational overhead. This combination of geometric flexibility and computational efficiency makes them particularly attractive for real-time AR/VR applications and resource-constrained robotic systems.

Point cloud generation has progressed through several modeling paradigms, from early VAEs \cite{kim2021setvae} and GANs \cite{li2018pointcloudgan, pmlr-v80-achlioptas18a}, which are fast at inference but often limited in fidelity and diversity and prone to mode collapse, to flow-based likelihood models such as PointFlow \cite{yang2019pointflow}, which provide tractable likelihoods via continuous normalizing flows but rely on ODE-solver-based iterative integration at sampling time. More recently, diffusion models, e.g., PVD \cite{zhou20213d} and LION \cite{zeng2022lion}, have become a dominant approach for high-quality 3D point cloud generation, though standard sampling can require hundreds to thousands of denoising steps, leading to significant latency and compute overhead. To address these costs, recent work has shifted toward few-step and single-step formulations, including Flow Matching, which reduces generation to learning a velocity field via regression \cite{lipman2023flow}, and sampling-path simplification methods such as Point Straight Flow (PSF) \cite{10203334}. Related flow-matching approaches such as NSOT \cite{hui2025nsot} leverage approximate optimal-transport couplings and hybrid training pairs to ease learning, partly motivated by challenges induced by straightened trajectories. Latest works include teacher-free continuous-time consistency models such as ConTiCoM-3D \cite{eilermann2025conticom3d} that target one- to two-step generation, as well as multi-scale flow-matching frameworks such as MFM-Point  \cite{molodyk2025mfmpointmultiscaleflowmatching} that improve scalability through geometry-aware cross-resolution alignment and coarse-to-fine generation.

Reducing the number of generative steps required by diffusion models has recently attracted considerable attention. Among the proposed approaches, MeanFlow \cite{geng2025meanflows} offers a principled framework for one-step generation by learning an average velocity field rather than the instantaneous velocity used in standard flow matching. Its training objective follows from the MeanFlow Identity, which establishes an intrinsic relationship between average and instantaneous velocities and provides a direct supervision signal for the neural network. MeanFlow requires neither pretrained models nor distillation or curriculum learning, yet achieves state-of-the-art image-generation quality using a single function evaluation (1-NFE). MeanFlow incorporates classifier-free guidance into the target velocity field, improving sample quality without additional inference-time computation.

Despite this promise, applying MeanFlow to point clouds raises a key challenge. In the extreme low-step regime, the model has little opportunity to correct errors along a long sampling trajectory, so generation quality depends strongly on the geometry of the training pairs. Optimal transport (OT) provides a natural way to construct geometry-consistent correspondences, but a naive OT formulation for point clouds is computationally expensive: training must account for both a shape-level coupling between source samples and target point clouds and a point-level coupling induced by the permutation invariance of points within each shape. Solving these nested matching problems directly makes OT-based training difficult to scale to dense point clouds.

Building on this observation, we propose \textbf{OT-MeanFlow3D}, a scalable one-step generative framework for point cloud generation and completion. Our key idea is to decouple global shape sampling from local point-level matching. Instead of solving an expensive outer OT problem between a batch of Gaussian samples and a batch of target shapes, we condition a Shape Generator on a shape-specific latent context. This reduces training to point-level OT within each conditioned shape, enabling geometry-consistent MeanFlow targets while avoiding the costly nested OT computation in point-cloud space. For unconditional generation, a lightweight Context Generator learns to sample the latent context distribution; for shape completion, the context is obtained by encoding the partial input. Thus, OT-MF3D retains the benefits of OT-based alignment while preserving efficient training and single-step inference.

We evaluate OT-MF3D on point cloud generation and completion, demonstrating improved speed--quality tradeoffs compared with prior diffusion-, flow-, and consistency-based methods. Our specific contributions are:

\begin{itemize}
\item We propose a scalable OT-MeanFlow framework for point clouds that combines MeanFlow's average-velocity objective with geometry-consistent point-level OT matchings, enabling high-quality one-step generation while avoiding expensive nested OT computation in point-cloud space through latent-space shape transport.

\item We introduce a latent-conditioned training framework for shape generation and completion, where a Context Generator models the latent shape distribution for unconditional generation and an encoder provides the context for completion.

\item Our approach achieves a strong speed--quality tradeoff, delivering competitive or state-of-the-art results in point cloud generation and completion while maintaining efficient training and single-step sampling.

\end{itemize}

\section{Related Work}
\label{sec:related}

\subsection{Point Cloud Generation}
Prior to diffusion models, point-cloud generative modeling was driven by latent-variable, adversarial, autoregressive, and likelihood-based approaches. Early work studied point-cloud autoencoders and latent/raw-space generators with fidelity and coverage metrics \cite{pmlr-v80-achlioptas18a}, while TreeGAN introduced tree-structured graph convolutions for multi-class synthesis \cite{shu20193d}. Autoregressive models such as PointGrow improve expressivity but sample points sequentially \cite{Sun_2020_WACV}, and likelihood-based models such as PointFlow provide tractable likelihoods through continuous normalizing flows but require iterative ODE sampling \cite{yang2019pointflow}. More recently, diffusion models \cite{ho2020denoising,song2021scorebased} have become a dominant paradigm for high-fidelity 3D generation and completion, including PVD \cite{zhou20213d} and LION \cite{zeng2022lion}, but their reliance on many denoising steps motivates faster alternatives.

Recent work has therefore focused on few-step and one-step point-cloud generation. PSF uses rectified flows to straighten trajectories for fast sampling \cite{10203334}, while NSOT constructs approximate OT-based training pairs and hybrid couplings to ease flow-matching training \cite{hui2025nsot}. Consistency-style objectives \cite{10.5555/3618408.3619743} have also been adapted to point clouds through ConTiCoM-3D, which targets one- to two-step generation without teacher distillation \cite{eilermann2025conticom3d}. Orthogonally, MFM-Point improves scalability with a coarse-to-fine multi-scale flow-matching design \cite{molodyk2025mfmpointmultiscaleflowmatching}. These methods suggest that in the low-step regime, the geometry of the transport path and the choice of pairing mechanism are central to generation quality.

\subsection{Coupling and Geometry in Low-Step Generation: Optimal Transport and Flow Matching}
\label{sec:rw_coupling_geometry}

Flow Matching (FM) learns a velocity field along a probability path from source to data \cite{lipman2023flow}. Because independent source--target couplings can yield crossing or curved trajectories that hinder low-NFE generation, recent methods use geometry-aware mini-batch OT couplings, often computed via Sinkhorn \cite{cuturi2013sinkhorn,tong2024minibatchot,pooladian2023multisample}.

Several recent works further study how coupling quality affects low-step generation. Optimal Flow Matching seeks straight OT-like displacements through restricted vector-field parameterizations \cite{kornilov2024ofm}. Rectified Flow and Simple ReFlow refine data--noise pairings through reflow-style training \cite{liu2023rectifiedflow,kim2025simplereflow}. LOOM-CFM improves minibatch coupling coherence by exchanging assignments across batches \cite{davtyan2025loom}, while Model-Aligned Coupling selects couplings based on model error rather than only geometric cost \cite{lin2025mac}. In the point-cloud domain, NSOT shows that fully OT-straightened training is challenging at scale and uses offline approximate OT and hybrid couplings to retain some benefits of geometry-consistent matching \cite{hui2025nsot}. Our work instead avoids the expensive nested OT in point-cloud space by using shape-level transport in a compact latent space.

\subsection{MeanFlow and One-Step Generation}
\label{sec:rw_meanflow}

MeanFlow \cite{geng2025meanflows} is a recent one-step generative framework that learns a time-averaged velocity field over finite intervals, in place of the instantaneous velocity field of standard flow matching. The identity relating the two fields yields a self-contained training objective, allowing one-step models to be trained from scratch without knowledge distillation or a pretrained teacher. A growing body of follow-up work targets the optimization and stability of this objective: Improved Mean Flows \cite{geng2025improved} reformulates the loss and introduces flexible guidance conditioning; Modular MeanFlow \cite{you2025modularmeanflow} develops stabilized loss formulations; and AlphaFlow \cite{zhang2025alphaflow} traces slow convergence to gradient conflicts and mitigates them through curriculum strategies. A complementary line of work couples mean-velocity learning with trajectory rectification or refinement, including Rectified MeanFlow \cite{zhang2025rectifiedmeanflow}, Pixel Mean Flows \cite{lu2026pmf}, and RMFlow \cite{huang2026rmflow}.

Prior work has largely pursued these two directions independently: improved couplings for flow matching on one hand, and few-step objectives such as MeanFlow on the other. How OT-based coupling geometry interacts with mean-velocity training in the extreme low-step regime remains largely unexplored. The question is particularly consequential for 3D point clouds, where inference efficiency and permutation-aware geometry are both central concerns. In this work, we show that combining point-level OT alignment with latent-conditioned MeanFlow enables efficient one-step point cloud generation and completion.

\section{Method}

\begin{figure*}[tb]
  \centering  
  \includegraphics[width=\textwidth]{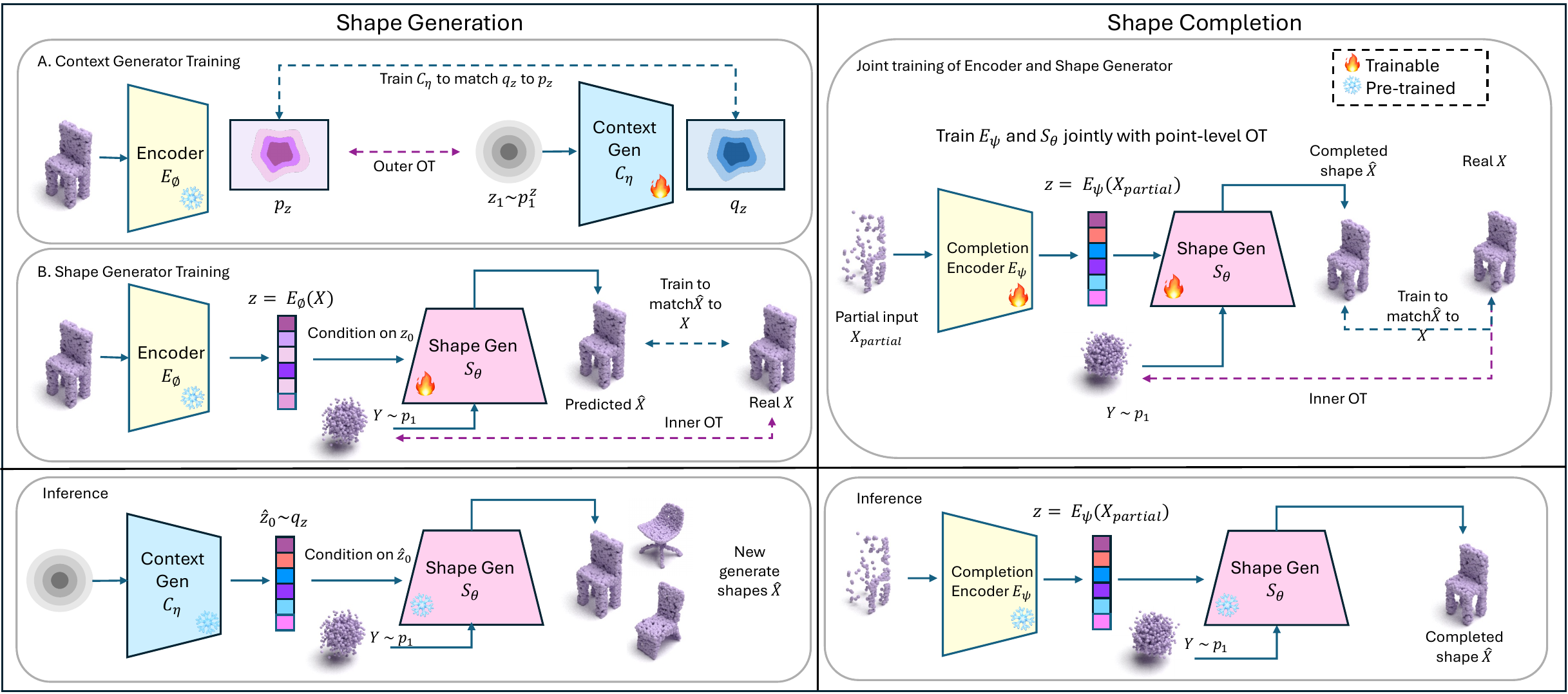}
  \caption{The overall framework for our training and inference process for shape generation and completion. \textbf{Shape Generation:} A pre-trained Encoder is used to train the Shape Gen and Context Gen, and Context Gen is used during inference to condition the Shape Gen model and generate unseen samples. \textbf{Shape completion:} An Encoder is trained jointly and the latent vector is used as conditioning.
  }
  \label{fig:overview}  
\end{figure*}

In this section, we present our method for efficient point-cloud generation and completion. Figure~\ref{fig:overview} provides an overview of the full system. We first introduce the MeanFlow and improved MeanFlow backbones used in our experiments. We then describe the computational challenge of applying optimal transport directly to point clouds and show how we decouple shape-level and point-level transport across latent and point-cloud representations. Finally, we describe the latent-conditioned training objective and the latent-space transport used for unconditional generation and completion.

\subsection{MeanFlow and Improved MeanFlow Backbone}
\label{sec:imf}

We build on flow matching (FM)~\cite{liu2023rectifiedflow,lipman2023flow,lipman2024flow}, MeanFlow (MF)~\cite{geng2025meanflows}, and improved MeanFlow (iMF)~\cite{geng2025improved}. Let $x_0\sim p_0$ denote a target data sample and $x_1\sim p_1$ a sample from a simple source distribution. We use the standard linear path
\begin{equation}
    z_t=(1-t)x_0+t x_1,
    \qquad v_c=x_1-x_0,
    \label{eq:imf_path}
\end{equation}
where $t=0$ corresponds to the data distribution and $t=1$ to the source distribution. Classical FM learns the instantaneous velocity along this path and typically requires multiple model evaluations to integrate the reverse-time ODE.

MeanFlow models interval-averaged velocity $[r,t]$~\cite{geng2025meanflows},
\begin{equation}
u(z_t,r,t)
=
\frac{1}{t-r}
\int_r^t v(z_\tau,\tau)\,d\tau,
    \label{eq:mean_velocity}
\end{equation}
which directly parameterizes the displacement between the interval endpoints and enables one-step generation. Differentiating this relation gives
\begin{equation}
u(z_t,r,t)
=
v(z_t,t)
-
(t-r)\frac{d}{dt}u(z_t,r,t).
    \label{eq:mf_identity}
\end{equation}
Original MF trains $u_\theta$ to regress the right-hand side of Eq.~\eqref{eq:mf_identity}, with the derivative target evaluated using the current network and stop-gradient. Specifically, defining
\begin{equation}
    U_\theta(z_t,r,t)
    =v_c-(t-r)\,\operatorname{JVP}
    \!\left(u_\theta;v_c\right),
    \label{eq:mf_target}
\end{equation}
the original MF objective can be written as
\begin{equation}
    \mathcal{L}_{\mathrm{MF}}
    =\mathbb{E}_{x_0,x_1,r,t}
    \!\left[
    \left\|u_\theta(z_t,r,t)
    -\operatorname{sg}\!\left(U_\theta(z_t,r,t)\right)
    \right\|_2^2\right],
    \label{eq:mf_loss}
\end{equation}
where the JVP evaluates the total derivative of $u_\theta$ along the path with tangent $(v_c,0,1)$.

Our model uses improved MeanFlow (iMF)~\cite{geng2025improved}, which rewrites Eq.~\eqref{eq:mf_identity} as instantaneous-velocity regression:
\begin{equation}
v(z_t,t)
=
u(z_t,r,t)
+
(t-r)\frac{d}{dt}u(z_t,r,t).
    \label{eq:imf_identity}
\end{equation}
thereby avoiding the network-dependent regression target of the original MF objective. Following the parameter-free boundary formulation of iMF, we estimate the instantaneous velocity inside the total derivative as
$v_\theta(z_t,t)=u_\theta(z_t,t,t)$ and define
\begin{equation}
V_\theta(z_t,r,t)
=u_\theta(z_t,r,t) +(t-r)\,\operatorname{JVP}_{\mathrm{sg}}
    \!\left(u_\theta;v_\theta\right),
\label{eq:imf_parameterization}
\end{equation}
where $\operatorname{JVP}_{\mathrm{sg}}$ denotes the total derivative of $u_\theta$ along the trajectory, evaluated with tangent $(v_\theta,0,1)$, with stop-gradient applied to the resulting JVP. The iMF objective is
\begin{equation}
    \mathcal{L}_{\mathrm{iMF}}
    =\mathbb{E}_{x_0,x_1,r,t}
    \!\left[\left\|V_\theta(z_t,r,t)-v_c\right\|_2^2\right].
    \label{eq:imf_loss}
\end{equation}

At inference, a single full-interval evaluation gives
$\hat{x}_0=x_1-u_\theta(x_1,0,1)$. Unless otherwise stated, we use iMF as our backbone; our contribution lies in how $x_0$ and $x_1$ are coupled for point-cloud training, described next.

\subsection{Optimal Transport for Point Clouds}
\label{sec:pc_ot}

Optimal transport has been used to improve flow matching by replacing independent source--target pairings with couplings that favor shorter transport paths~\cite{pooladian2023multisample,tong2024minibatchot}. For point clouds, applying this idea directly introduces an additional matching problem. A point cloud is an unordered set,
\[
X=\{x_i\}_{i=1}^{N},
\qquad x_i\in\mathbb{R}^{3},
\]
so comparing two point clouds requires accounting for both \emph{which shapes are paired} and \emph{which points within those shapes correspond}.

Consider a minibatch of source clouds
$\{Y_b\}_{b=1}^{B}$ and target point clouds
$\{X_c\}_{c=1}^{B}$, each containing $N$ points. A permutation-aware cost between a candidate source--target pair can be written as
\[
C_{bc}
=
\min_{\gamma\in\Pi_N}
\sum_{i,j}
\gamma_{ij}
\|y_{b,i}-x_{c,j}\|_2^2,
\]
where $\Pi_N$ denotes the set of admissible point-level couplings. An outer mini-batch OT problem can then use the costs $C_{bc}$ to determine which source cloud should be paired with each target cloud.

This yields nested transport: an \emph{outer} coupling between clouds and an \emph{inner} pointwise alignment for each candidate pair. Exact matching computes these alignments throughout the outer cost matrix, costing $O(B^2N^3)$ and making mini-batch OT impractical for dense clouds~\cite{hui2025nsot}.

We decouple these levels: outer transport operates in latent space without pointwise alignment, while permutation-aware transport is applied within each conditioned target shape. The next section describes the latter.

\subsection{Latent-Conditioned Point-Level OT}
\label{sec:latent_ot}

We decouple the nested transport problem by representing global shape variation in a compact latent space. During training, the flow model is conditioned on a latent \(z\) representing either the target point cloud for generation or the partial point cloud for completion. This removes the need to solve shape-level OT over full point clouds when constructing point-level training pairs.

Given a target point cloud
$X=\{x_j\}_{j=1}^{N}$, we use $z=E_\phi(X)$ for generation and
$z=E_\psi(\widetilde{X})$ for completion.
We independently sample a source cloud
$Y=\{y_i\}_{i=1}^{N}$ and compute only the point-level OT coupling
\begin{equation}
\gamma^\star
=
\arg\min_{\gamma\in\Pi_N}
\sum_{i,j}\gamma_{ij}
\|y_i-x_j\|_2^2 .
\label{eq:point_ot}
\end{equation}
Unlike mini-batch OT-CFM~\cite{pooladian2023multisample,tong2024minibatchot}, Eq.~\eqref{eq:point_ot} does not couple different shapes within a mini-batch. Only one source cloud is aligned with each conditioned target shape, reducing the nested $O(B^2N^3)$ construction of Sec.~\ref{sec:pc_ot} to $B$ independent point-level matching problems.

Additionally, this construction allows us to further reduce the cost of point-level OT by randomly subsampling \(P<N\) points from each point cloud at every training iteration and solving Eq.~\eqref{eq:point_ot} on these subsets. Since global shape information is provided through the latent condition \(z\), the flow model need not process all points simultaneously. Different subsets are observed across training iterations while OT is solved only over \(P\) points.

For a matched pair $(y_i,x_j)\sim\gamma^\star$, the flow path is
\begin{equation}
x_{t,ij}=(1-t)x_j+t y_i,
\qquad
v_{ij}=y_i-x_j .
\label{eq:ot_path}
\end{equation}
We then condition the Shape Generator on \(z\) and train with one of the MF objectives defined in Sec. \ref{sec:imf}. For original MF, define the regression target as
\begin{equation}
U_\theta^{\mathrm{OT}}(x_{t,ij},r,t,z)=
v_{ij}-(t-r)\,\operatorname{JVP}
\!\left(u_\theta;v_{ij}\right).
\label{eq:ot_mf_target}
\end{equation}
We then optimize
\begin{equation}
\mathcal{L}_{\mathrm{OT\text{-}MF}}
=\mathbb{E}\!\left[
\left\|u_\theta(x_{t,ij},r,t,z)
-\operatorname{sg}\!\left(U_\theta^{\mathrm{OT}}\right)
\right\|_2^2\right].
\label{eq:ot_mf}
\end{equation}
Likewise for improved MeanFlow:
\begin{equation}
\mathcal{L}_{\mathrm{OT\text{-}iMF}}
=
\mathbb{E}
\left[
\left\|
V_\theta(x_{t,ij},r,t,z)-v_{ij}
\right\|_2^2
\right],
\label{eq:ot_imf}
\end{equation}
where the expectations are over target shapes, source clouds, point pairs sampled from $\gamma^\star$, and the corresponding MF/iMF time variables. In both variants, optimal transport changes only the coupling used to construct source--target point pairs; the underlying MF or iMF objective is otherwise unchanged.

This decomposition suits point clouds: transport in a compact latent space captures global shape, while point-level OT preserves permutation-aware correspondence within each conditioned shape.

\subsection{Latent-Space Shape Transport}
\label{sec:latent_shape_ot}

The latent-conditioned Shape Generator in Sec.~\ref{sec:latent_ot} avoids outer transport in point-cloud space; for unconditional generation, we instead perform global shape transport in the encoder’s compact latent space. This decouples the nested problem into latent shape-level transport and permutation-aware point-level transport within each conditioned shape. We denote the distribution of encoded target latents $z_0=E_\phi(X)$ by $p_z$ and the distribution of generated latents $\hat z_0=C_\eta(z_1)$ by $q_z$.

\noindent\textbf{Unconditional generation.}
We first train a lightweight autoencoder $(E_\phi,D_\phi)$ on complete training shapes. Each target point cloud $X$ is mapped to a latent representation
    $z_0 = E_\phi(X)$,
which defines the target latent distribution. During Shape Generator training, $E_\phi$ is frozen and $z_0$ is used as the conditioning input in Eq.~\eqref{eq:ot_imf}.

To generate shape latents at inference time, we train a Context Generator to transport samples from a simple latent source distribution to the encoded target distribution. Given source latent samples $z_1\sim p_1^z$, we construct mini-batch OT couplings between source and target latents. For source and target mini-batches
$\{z_1^{(b)}\}_{b=1}^{B}$ and
$\{z_0^{(c)}\}_{c=1}^{B}$, respectively, we solve, over couplings $\Pi_B$ between the two mini-batches,
\begin{equation}
    \pi_z^\star
    =
    \arg\min_{\pi\in\Pi_B}
    \sum_{b,c}
    \pi_{bc}
    \|z_1^{(b)}-z_0^{(c)}\|_2^2 .
    \label{eq:latent_ot}
\end{equation}
Unlike shape-level OT in the original point-cloud space, each entry of this cost matrix is a direct latent-space distance and therefore does not require an inner point-matching problem. The Context Generator $C_{\eta}$ is then trained with the same MF/iMF formulation from Sec.~\ref{sec:imf} using pairs sampled from $\pi_z^\star$.

At inference, a source latent is transported by the Context Generator to obtain a shape condition $z$. A source point cloud is then transformed by the Shape Generator conditioned on $z$, yielding the generated point cloud.

\noindent\textbf{Shape completion.}
For completion, the desired global shape information is partially observed and does not need to be generated from the latent source distribution. Instead, a partial-shape encoder $E_\psi$ maps the observed point cloud $\widetilde{X}$ directly to the conditioning latent,
    $z = E_\psi(\widetilde{X})$.
$E_\psi$ and the Shape Generator are trained jointly using Eq.~\eqref{eq:ot_imf}, with the OT coupling computed between a source point cloud and the corresponding complete target shape. Because the partial observation directly provides the global context, the latent-space Context Generator and its outer OT problem are bypassed during completion.

For unconditional generation, our formulation uses global and local transport across latent and point spaces, avoiding their expensive nested computation in point-cloud space. For completion, the observed partial shape directly provides the global context, so only the conditioned point-level transport is required.
Algorithm~\ref{alg:latent-cond-otmf} shows our full training and inference pipeline for shape generation and completion.

\begin{algorithm}[t]
\caption{OT-MF3D Training/Inference}
\label{alg:latent-cond-otmf}
\small
\begin{algorithmic}[1]
\Require Target shapes $X\sim\mathcal{D}_{\mathrm{data}}$, source distribution $p_1$, partial/full pairs $(\widetilde{X},X)$
\Ensure Context Generator $C_\eta$, Shape Generator $S_\theta$
\Statex \textbf{Context Generator training (generation only)}
\State Train autoencoder $(E_\phi,D_\phi)$ on $X\sim\mathcal{D}_{\mathrm{data}}$; freeze $E_\phi$ and set $z_0=E_\phi(X)$.
\For{each iteration}
    \State Sample $z_1\sim p_1^z$; compute mini-batch OT coupling between $z_1$ and $z_0$.
    \State Update $C_\eta$ with the MF/iMF objective on OT-coupled latent pairs.
\EndFor
\Statex \textbf{Shape Generator training}
\For{each iteration}
    \State Obtain context: $z=E_\phi(X)$ (generation) or $z=E_\psi(\widetilde{X})$ (completion).
    \State Sample $Y\sim p_1$; compute point-level OT coupling $\gamma^\star$ between $Y$ and $X$.
    \State Update $S_\theta$ with $\mathcal{L}_{\mathrm{OT\text{-}MF}}$ or $\mathcal{L}_{\mathrm{OT\text{-}iMF}}$ on $\gamma^\star$-matched pairs (jointly with $E_\psi$ for completion).
\EndFor
\Statex \textbf{Inference}
\State For generation, sample $z_1\sim p_1^z$ and set
$z=C_\eta(z_1)\equiv z_1-u_\eta^z(z_1,0,1)$. For completion, set $z=E_\psi(\widetilde{X})$.
\State Sample $Y\sim p_1$ and output
$\widehat{X}=S_\theta(Y,z)\equiv Y-u_\theta(Y,0,1,z)$.
\end{algorithmic}
\end{algorithm}

\section{Experiments}
In this section, we describe the datasets used for evaluation and present our results on both unconditional point cloud generation and conditional point cloud completion.

\subsection{Unconditional Generation}

\noindent\textbf{Datasets and Metrics.}
Following prior work on unconditional shape generation \cite{hui2025nsot, yang2019pointflow, 10203334, zhou20213d}, we utilize the ShapeNet \cite{chang2015shapenetinformationrich3dmodel} Chair, Car, and Airplane categories. We adopt the same preprocessing pipeline as previous methods \cite{10203334, zhou20213d}. Specifically, we use the same train/test split and sample 2,048 points from each point cloud in the dataset. As in prior work, we train a separate generative model for each category. Additionally, we train a model on all classes of ModelNet-10~\cite{wu20153d} and show qualitative results.

For evaluation, we report the one-nearest-neighbor accuracy (1-NNA) using Chamfer Distance (CD) and Earth Mover’s Distance (EMD), following standard practice in unconditional shape generation \cite{hui2025nsot, 10203334, yang2019pointflow}. For this metric, values closer to $50\%$ indicate better generation quality. In addition, we report Sampling Time (ST), measured in seconds. Our inference time measurement closely follows PSF \cite{10203334}. Specifically, we measure the sampling time with a batch size of 1 and average the results over 50 random trials on a single Nvidia RTX A6000 GPU. For fast methods (ST $<$ 0.04), which have a similar ST to ours, we measure ST on our GPU as well to ensure a fair comparison. These methods include 1-GAN~\cite{pmlr-v80-achlioptas18a}, SetVAE~\cite{kim2021setvae}, and PSF~\cite{10203334}.

\noindent\textbf{Baselines.}
We compare our method with baselines reported in PSF~\cite{10203334}, including PointFlow~\cite{yang2019pointflow}, DPF-Net~\cite{10.1007/978-3-030-58592-1_41}, SoftFlow~\cite{10.5555/3495724.3497099}, ShapeGF~\cite{10.1007/978-3-030-58580-8_22}, DPM~\cite{luo2021diffusion}, 1-GAN~\cite{pmlr-v80-achlioptas18a}, SetVAE~\cite{kim2021setvae}, and PVD~\cite{zhou20213d}. Results for these baselines are taken from ~\cite{10203334}, as we use the same datasets and evaluation metrics. For PSF\cite{10203334}, we utilized their published code and trained their model on SN-Chair and Airplane. We also include recent methods such as LION~\cite{zeng2022lion}, NSOT~\cite{hui2025nsot}, and ConTiCoM-3D~\cite{eilermann2025conticom3d}. ConTiCoM-3D is the most recent one-step point cloud generation method.

\begin{figure}[t]
  \centering
   \includegraphics[width=\linewidth]{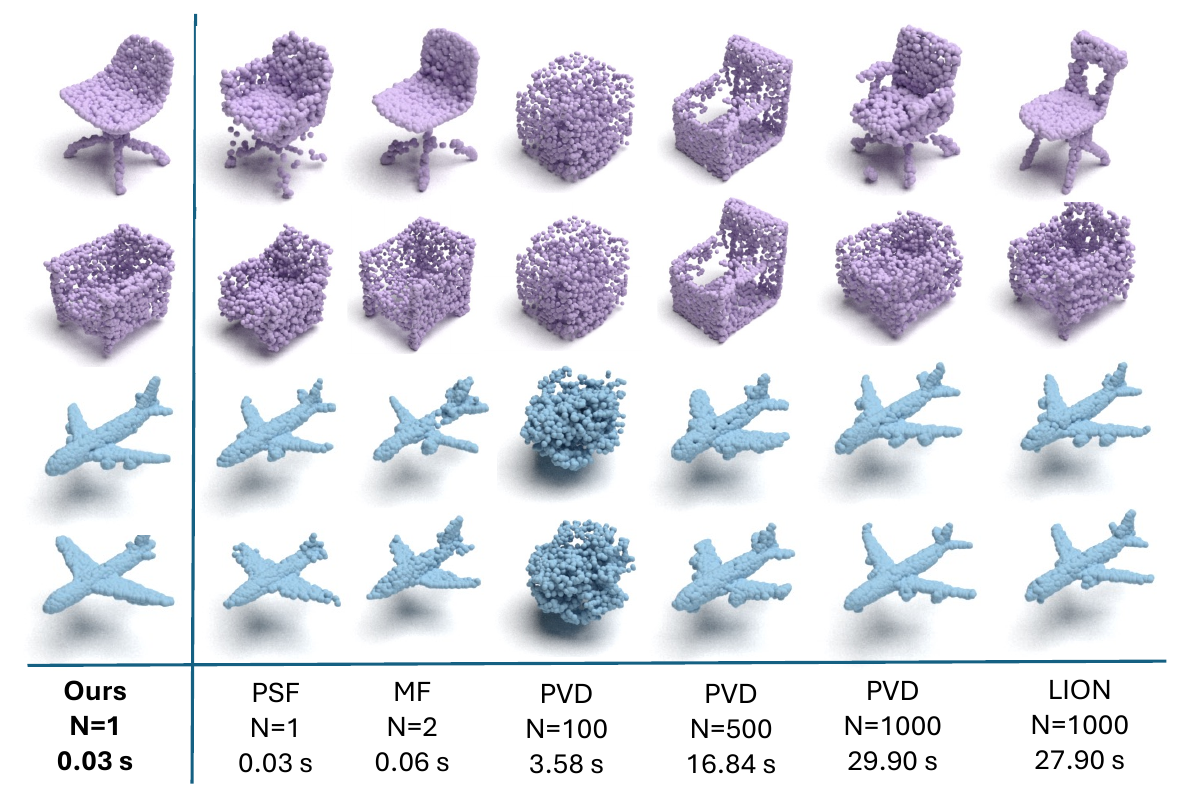}
\vspace{-2.\baselineskip}
   \caption{Qualitative comparison. Times measured on our GPUs.}
   \vspace{-.5\baselineskip}
   \label{fig:qual}
\end{figure}

\begin{table*}[t]
\caption{ShapeNet generation results. For 1-NN-CD/EMD, values closer to 50 are better. Bottom-block methods have inference times comparable to our one-step model. \textbf{Bold} and \underline{underlined} denote the best and second-best results per block. iMF is denoted with (imp). Our method ranks first or second among the fastest methods while training more efficiently than PSF~\cite{10203334}. PSF* refers to our results generated using the public code and training recipe.}
\vspace{-.5\baselineskip}
\label{tab:shapenet_generation}
\centering
\footnotesize
\setlength{\tabcolsep}{2pt}
\begin{threeparttable}
\begin{tabular*}{\textwidth}{@{\extracolsep{\fill}} l c cc cc cc cc @{}}
\toprule
\multirow{2}{*}{Model} & 
\multirow{2}{*}{ST (s)} & 
\multicolumn{2}{c}{Airplane} & 
\multicolumn{2}{c}{Chair} & 
\multicolumn{2}{c}{Car} & 
\multicolumn{2}{c}{Avg} \\
\cmidrule(lr){3-4}
\cmidrule(lr){5-6}
\cmidrule(lr){7-8}
\cmidrule(lr){9-10}
& & CD & EMD
    & CD & EMD
    & CD & EMD
    & CD & EMD \\
\midrule
PointFlow\cite{yang2019pointflow}           & 0.27 & 75.68 & 70.74 & 62.84 & 60.57 & 58.10 & 56.25 & 65.54 & 62.52 \\
DPF-Net\cite{10.1007/978-3-030-58592-1_41}  & 0.33 & 75.18 & 65.55 & 62.00 & 58.53 & 62.35 & 54.48 & 66.51 & 59.52 \\
SoftFlow\cite{10.5555/3495724.3497099}      & \textbf{0.12} & 76.05 & 65.80 & 59.21 & 60.05 & 64.77 & 60.09 & 66.68 & 61.98 \\
ShapeGF\cite{10.1007/978-3-030-58580-8_22}  & 0.34 & 80.00 & 76.17 & 68.96 & 65.48 & 63.20 & 56.53 & 70.72 & 66.06 \\
DPM\cite{luo2021diffusion}                  & 22.8 & 76.42 & 86.91 & 60.05 & 74.77 & 68.89 & 79.97 & 68.45 & 80.55 \\
PVD \cite{zhou20213d}(N=1000)               & 29.9 & 73.82 & 64.81 & 56.26 & \underline{53.32} & 54.55 & 53.83 & 61.54 & 57.32 \\
PVD-DDIM \cite{song2021denoising}(N=100)    & 3.15 & 76.21 & 69.84 & 61.54 & 57.73 & 60.95 & 59.35 & 66.23 & 62.31 \\
LION \cite{zeng2022lion}(N=1000)            & 27.09 & \underline{67.41} & \textbf{61.23} & \textbf{53.70} & 53.41 & \underline{53.41} & \textbf{51.14} & \underline{58.17} & \underline{55.26} \\
NSOT \cite{hui2025nsot}(N=1000)             & N/A & 68.64 & 61.85 & 55.51 & 57.63 & 59.66 & 53.55 & 61.27 & 57.68 \\
ConTiCoM-3D\cite{eilermann2025conticom3d} (N=1) & \underline{0.22} & \textbf{64.89} & \underline{61.77} & \underline{54.30} & \textbf{50.52} & \textbf{53.30} & \underline{51.40} & \textbf{57.50} & \textbf{54.56} \\
\midrule
1-GAN\cite{pmlr-v80-achlioptas18a}          & \textbf{0.001} & 87.30 & 93.95 & 68.58 & 83.84 & 66.49 & 88.78 & 74.12 & 88.86 \\
SetVAE\cite{kim2021setvae}                  & \underline{0.03} & 75.31 & 77.65 & 58.76 & 61.48 & 59.66 & 61.48 & 64.58 & 66.87 \\
PSF \cite{10203334}                         & \underline{0.03} & \textbf{71.11} & \textbf{61.09} & 58.92 & 54.45 & \textbf{57.19} & 56.07 & \textbf{62.41} & \underline{57.20} \\
PSF* \cite{10203334}                         & \underline{0.03} & 85.43 & 84.94 & 59.74 & 62.24 & -- & -- & -- & -- \\
\rowcolor{gray!15}
\textbf{OT-MF3D (Ours)} (N=1)               & \underline{0.03} & 72.71 & \underline{62.22} & \underline{57.77} & \textbf{52.79} & 60.79 & \underline{54.54} & 63.76 & \textbf{56.52} \\
\rowcolor{gray!15}
\textbf{OT-MF3D (imp) (Ours)} (N=1)         & \underline{0.03} & \underline{71.60} & 64.07 & \textbf{57.40} & \underline{54.08} & \underline{58.80} & \textbf{53.97} & \underline{62.60} & 57.37 \\
\bottomrule
\end{tabular*}
\end{threeparttable}
\end{table*}

\begin{figure*}[t!]
  \centering  
  \vspace{-1.\baselineskip}
   \includegraphics[width=\textwidth]{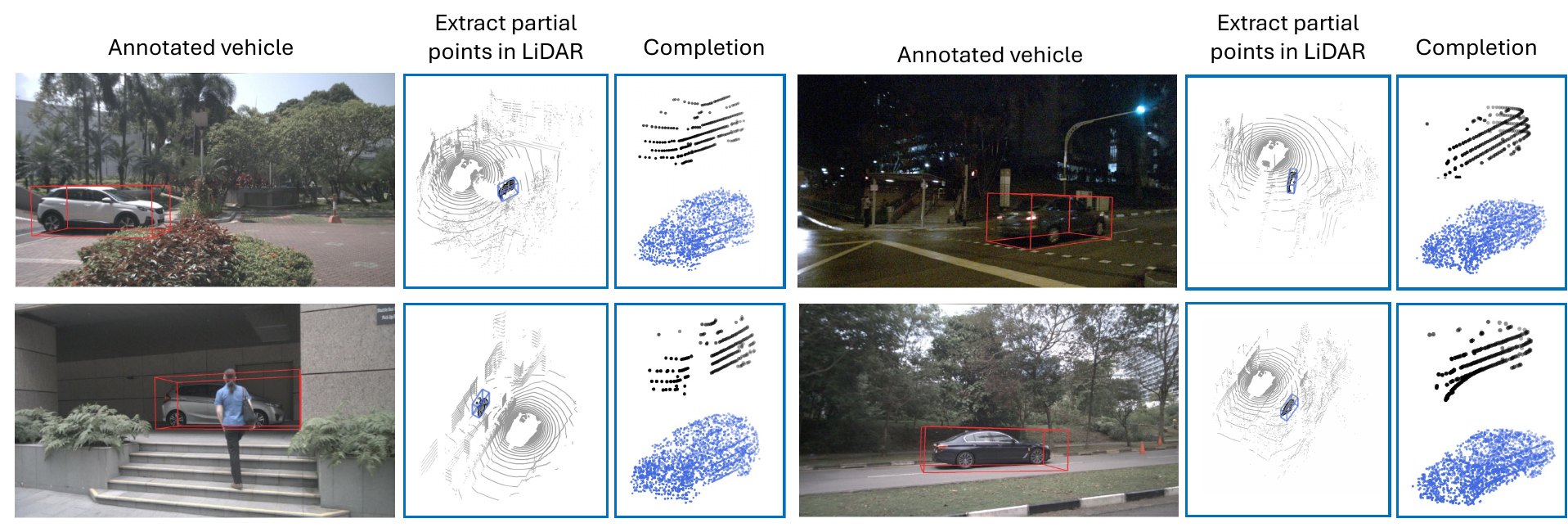}
   \vspace{-2.\baselineskip}
  \caption{Completion samples using our one-step completion model on the nuScenes dataset.
  }
  \vspace{-.5\baselineskip}
  \label{fig:completion_lidar}  
\end{figure*}

\noindent\textbf{Results.}
As shown in Table~\ref{tab:shapenet_generation}, our method achieves best or second-best results among one-step baselines with comparable inference time, such as PSF~\cite{10203334} and SetVAE~\cite{kim2021setvae}. ConTiCoM-3D~\cite{eilermann2025conticom3d}, while also one-step, is more than $5\times$ slower yet only comparable in 1-NNA metrics.
Our method is also cheaper to train. Table~\ref{tab:training_eff} reports training time (TT, hours) and peak GPU memory against PVD~\cite{zhou20213d}, LION~\cite{zeng2022lion}, ConTiCoM-3D~\cite{eilermann2025conticom3d}, and PSF~\cite{10203334}, each run with its original best-performing settings measured on our GPUs. Ours achieves the best training time and second-best memory usage.

\begin{table}[t]
\caption{Comparison of training time (TT) and peak GPU memory measured on our GPUs.}
\label{tab:training_eff}
\centering
\footnotesize
\setlength{\tabcolsep}{2pt}
\begin{tabular*}{\columnwidth}{@{\extracolsep{\fill}} l cc @{}}
\toprule
Method & TT (h) & Memory (GB) \\
\midrule
PVD~\cite{zhou20213d}                        & 107 & \textbf{3.78} \\
LION~\cite{zeng2022lion}                     & 273 & 32.1 \\
ConTiCoM-3D~\cite{eilermann2025conticom3d}   & 146 & 37.4 \\
PSF~\cite{10203334}                          & \underline{85} & 38.8 \\
\midrule
\rowcolor{gray!15}
\textbf{OT-MF3D (imp)(Ours)}                                & \textbf{31} & \underline{15.7} \\
\bottomrule
\end{tabular*}
\end{table}

Figure~\ref{fig:gen_samples} shows diverse, high-quality samples generated in under $0.04$ seconds across datasets. Figure~\ref{fig:qual} compares our outputs with the closest samples from PSF~\cite{10203334}, MF without OT, PVD~\cite{zhou20213d}, and LION~\cite{zeng2022lion}. For PSF~\cite{10203334}, we trained their model using the official code. For PVD~\cite{zhou20213d} and  LION~\cite{zeng2022lion}, we utilized the official checkpoints. NSOT~\cite{hui2025nsot} is omitted due to unavailable reproducibility details. Our method meets or improves quality with up to $747\times$ faster inference.

\subsection{Shape Completion}

To further evaluate the capabilities of our method, we conduct experiments on the point cloud completion task. In this setting, a subset of points is sampled from the full shape, and the model is expected to generate points to reconstruct the complete shape.

\noindent\textbf{Datasets and Metrics.}
We use the GenRe dataset introduced in \cite{10.5555/3327144.3327153}, following prior work. The dataset includes 20 random views for depth estimation. From each single-view observation, we sample 200 points and use their encoded representation as conditioning input to our model. Additional implementation details are provided in Appendix~A.
For evaluation, we report Earth Mover's Distance (EMD), as previous studies \cite{zhou20213d, 10203334} have shown that it is a more suitable metric for the shape completion task.

\noindent\textbf{Baselines.}
We compare our method with several existing approaches, including SoftFlow \cite{10.5555/3495724.3497099}, PointFlow \cite{yang2019pointflow}, DPF-Net \cite{10.1007/978-3-030-58592-1_41}, PVD \cite{zhou20213d}, and PSF \cite{10203334}. The reported numbers are taken from \cite{10203334}, and we use the same datasets, evaluation metrics, and inference time computation for a fair comparison. We measure the ST of PSF~\cite{10203334} on our GPU as well, since it is the only method with on-par inference time compared to ours. We exclude NSOT \cite{hui2025nsot}, as it evaluates shape completion only on the ShapeNet-Chairs.

\begin{figure}[t]
  \centering  
   \includegraphics[width=\columnwidth]{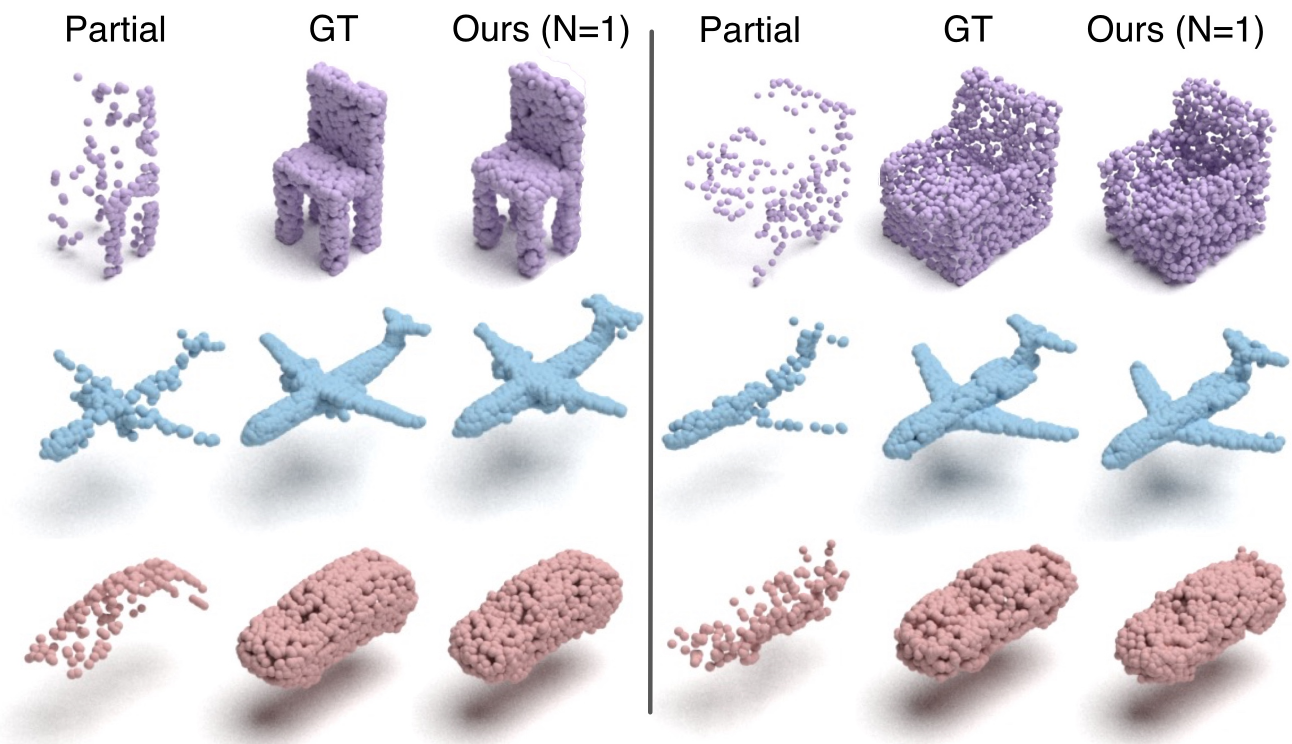}
   \vspace{-1.\baselineskip}
  \caption{Partial point cloud, ground truth, and completed samples using our one-step completion model. Our shape completion model can successfully capture the full shape from a partial view. 
  }
  \vspace{-.5\baselineskip}
  \label{fig:completion}  
\end{figure}

\noindent\textbf{Results.}
The results are summarized in Table \ref{tab:completion_res}. As shown in the table, our method achieves improved performance compared to the baselines, including PSF~\cite{10203334}, which is also a flow-based one-step generation method. In addition, our method attains the fastest runtime among the compared approaches, enabling efficient shape completion while maintaining high reconstruction quality. Qualitative completion results are also shown in Figure~\ref{fig:completion}.

\noindent\textbf{LiDAR data shape completion.}
To evaluate transfer beyond synthetic partial scans, we apply the frozen SN-Car-trained model, without fine-tuning, to real LiDAR sweeps from nuScenes~\cite{Caesar_2020_CVPR}. For each annotated vehicle, we crop points within its ground-truth 3D bounding box, transform them to a box-aligned canonical frame, and isotropically rescale them to match the mean car length used during training. We then apply the SN-Car normalization statistics and perform completion in a single sampling step.

We use ground-truth boxes to isolate shape completion from vehicle detection and localization errors. Thus, this experiment evaluates whether the learned shape prior transfers to the noise and sensor-specific sampling patterns of real LiDAR, rather than demonstrating an end-to-end perception pipeline.
Figure \ref{fig:completion_lidar} shows examples
across varying distances and occlusion levels. Despite the domain gap from synthetic training data, the model produces plausible complete car geometries, suggesting that the learned prior can generalize to real-world partial observations.

\begin{table}[t]
  \caption{Point-cloud completion quality and latency. EMD is scaled by $10^2$. Our method achieves the best EMD and fastest inference.}
  \vspace{-.5\baselineskip}
  \label{tab:completion_res}
  \centering
  \small
  \begin{tabular}{@{}lcccc@{}}
    \toprule
    & & \multicolumn{3}{c}{EMD $\downarrow$} \\
    \cmidrule(lr){3-5}
    Model & ST (s) 
    & Airplane & Chair & Car \\
    \midrule

    SoftFlow\cite{10.5555/3495724.3497099}
      & 0.12 & 1.198 & 3.295 & 2.789 \\

    PointFlow\cite{yang2019pointflow}
      & 0.27 & 1.180 & 3.649 & 2.851 \\

    DPF-Net\cite{10.1007/978-3-030-58592-1_41}
      & 0.34 & 1.105 & 3.320 & 2.318 \\

    PVD \cite{zhou20213d}(N=1000)
      & 29.98 & 1.030 & 2.939 & 2.146 \\

    PSF \cite{10203334}(N=1)
      & \textbf{0.03}
      & 1.004
      & 2.937
      & 2.194 \\
    \midrule
    \textbf{OT-MF3D(Ours)} (N=1) 
      & \textbf{0.03} & \textbf{0.51} & \textbf{1.71} & \textbf{1.4} \\

    \bottomrule
  \end{tabular}
\vspace{-.5\baselineskip}
\end{table}

\section{Ablations}
\noindent\textbf{Effect of inner OT batch size.}
In training the shape generator for OT-MF3D, the number of point clouds in a batch does not determine the quality of the OT assignment, since outer OT is not solved during training of the shape generator and the context generator is frozen. 
Instead, the relevant OT-related design choice is the number of points used for the per-shape point-level matching. We ablate this using the MF objective. Consistent with baselines, the initial number of points is $2,048$. During training, we subsample $P$ points for faster training. This does not affect the inference. During inference, the model generates the full $2,048$ shape. $B$ is the effective batch size.  Table~\ref{tab:batch_ablation} reports this ablation on ShapeNet-Chair. Reducing the point cloud batch size $B$ from $64$ to $32$ gives comparable performance, which is expected. Reducing the point-level OT size from $P=256$ to $P=128$ slightly degrades performance, as expected, but remains competitive among fast baselines. This ablation does not study the effect of outer OT batch size in the Context Generator, and specifically analyzes the inner OT batch size.

\begin{table}[t]
\caption{Effect of varying number of point clouds (B) vs number of subsampled points (P).}
\label{tab:batch_ablation}
\centering
\footnotesize
\setlength{\tabcolsep}{2pt}
\begin{tabular*}{\columnwidth}{@{\extracolsep{\fill}} cccc @{}}
\toprule
$B$ & $P$ & CD & EMD \\
\midrule
64 & 256 & 57.77 & \textbf{52.79} \\
32 & 256 & \textbf{57.17} & 53.77 \\
64 & 128 & 58.00 & 53.92 \\
\bottomrule
\end{tabular*}
\end{table}

\noindent\textbf{Effect of adding OT.} In Table~\ref{tab:chairs_abl}, we ablate the outer and inner OT components on ShapeNet-Chair, keeping all other settings fixed. For standard MeanFlow, inner OT substantially improves one-step generation, reducing CD/EMD from $97.43/93.50$ to $57.77/52.79$. In the improved MeanFlow (iMF) variant, inner OT again improves clearly over iMF alone while outer OT mainly improves CD. Neither alone is sufficient, and combining both is the most effective, indicating that the two OT components are complementary.

\newcommand{\cmark}{\ding{51}} %
\newcommand{\xmark}{\ding{55}} %

\begin{table}[tb]
\caption{1-NNA comparison on ShapeNet Chairs. Outer and Inner OT denote distribution- and point-level transport, respectively.}
\label{tab:chairs_abl}
\centering
\footnotesize
\setlength{\tabcolsep}{2pt}
\begin{tabular*}{\columnwidth}{@{\extracolsep{\fill}} lccccc @{}}
\toprule
Model & Outer & Inner & ST (s) & CD & EMD \\
\midrule
MF ($N=1$)        & \cmark & \xmark & 0.03 & 97.43 & 93.50 \\
MF ($N=2$)        & \cmark & \xmark & 0.06 & 62.46 & 63.14 \\
OT-MF3D ($N=1$)   & \cmark & \cmark & 0.03 & \textbf{57.77} & \textbf{52.79} \\
\midrule
MF (imp)          & \xmark & \xmark & 0.03 & 59.96 & 60.42 \\
OT-MF3D (imp)     & \xmark & \cmark & 0.03 & 59.36 & 55.43 \\
OT-MF3D (imp)     & \cmark & \xmark & 0.03 & 59.06 & 60.57 \\
OT-MF3D (imp)     & \cmark & \cmark & 0.03 & \textbf{57.40} & \textbf{54.08} \\
\bottomrule
\end{tabular*}
\vspace{-.5\baselineskip}
\end{table}

\section{Conclusion and Limitations}
We introduced OT-MF3D, which decouples global shape sampling from
point-level matching through latent conditioning, enabling exact
point-level OT within each conditioned shape while performing
shape-level transport in a compact latent space. On ShapeNet generation and completion, this improves the speed--quality tradeoff over diffusion-, flow-, and one-step baselines at a fraction of the training cost. We do not yet evaluate larger-scale 3D datasets, and leave extensions to other domains to future work.

\noindent\textbf{Acknowledgments.} This work was supported by NSF CAREER Award No. 2339898 and by a generous gift from Toyota Motor North America.

\clearpage
{
    \small
    \bibliographystyle{ieeenat_fullname}
    \bibliography{references}
}

\clearpage
\onecolumn
\begin{center}
  {\Large \bf Supplementary Material \par}
  \vspace*{12pt}
\end{center}
\appendix

\section{Additional Implementation Details}
\label{sec:impl}

\subsection{Architecture}
\label{sec:appendix-arch}

All OT-MF3D experiments use a pointwise residual MLP as the MeanFlow generator. The network consists of pre-activation residual blocks with LayerNorm, GELU, and linear layers, followed by a final output projection. For each point, the generator takes as input its current state, the MeanFlow time variables $t$ and $h=t-r$, and, when applicable, a shape-level conditioning vector shared across all points of the same shape.

For shape generation, the point-cloud generator is conditioned on a $512$-dimensional latent obtained from the pretrained autoencoder, or from the context generator. It uses a hidden dimension of $2048$ and $12$ residual blocks. The context generator uses the same residual-MLP design in the $512$-dimensional latent space, but without additional shape conditioning, with hidden dimension $1024$ and $4$ blocks.

For shape completion, the generator is conditioned on a $256$-dimensional representation of the partial input produced by a PVCNN2-based encoder following \citet{zhou20213d}. The encoder and generator are trained jointly end-to-end. The completion generator uses hidden dimension $2048$ and $12$ residual blocks.

The generator operates independently on each generated point; interactions between points are introduced only through the shared shape-level conditioning vector. No self-attention or explicit point-to-point communication is used within the generator.

\paragraph{Pointwise denoising with OT supervision.}
Our residual-MLP generator is intentionally pointwise: each generated point is processed independently given its current state, the MeanFlow time variables, and a shared global shape latent. Consequently, the generator is not explicitly set-aware and does not observe the other points being generated. In contrast, the OT coupling used to construct the training targets is computed jointly over the complete source and target point sets, so each pointwise target reflects the global configuration of the two sets.

 Although the network itself does not perform explicit point-to-point interaction, the OT assignment provides set-level structure in the supervision by determining globally consistent correspondences between source and target points. The shared shape latent further provides global information about the desired geometry. Thus, set-level information enters the learning problem through the OT targets and global conditioning, rather than through an explicitly set-aware denoising backbone.

This design also provides a favorable efficiency tradeoff. Several point-cloud generative baselines employ more set-aware architectures with explicit point interactions. In our main results, we compare with baselines such as PSF, PVD, NSOT, and SetVAE, where set-aware backbones are used. Such architectures are generally more complex and can incur substantially higher training and inference costs. In contrast, our pointwise MLP remains lightweight while achieving competitive generation quality and enabling efficient one-step sampling.

\subsection{MeanFlow Objective and Loss Normalization}
\label{sec:appendix-loss-and-mf}

Our OT-enhanced framework supports both MeanFlow (MF) and improved MeanFlow (iMF) training. In both cases, we follow the corresponding MeanFlow formulation and implementation of \cite{geng2025meanflows}, while replacing the independently sampled source--target pairing with our OT-based coupling. Thus, the MeanFlow objective and its loss normalization are adopted from the original method and are not specific contributions of OT-MF3D.

Following the original MeanFlow implementation, the per-example objective is normalized according to its detached loss magnitude:

$$
\mathrm{loss}
=
\frac{1}{B}\sum_{i=1}^{B}
\frac{\ell_i}
{\big(\mathrm{sg}[\ell_i]+\epsilon\big)^p},
$$

where $\ell_i$ denotes the MeanFlow regression loss for example $i$, $\mathrm{sg}[\cdot]$ denotes stop-gradient, and we use $\epsilon=10^{-6}$ and $p=0.75$ in all experiments. These settings follow the original MeanFlow codebase and are kept unchanged in our implementation.

\subsection{Shape Generation}

Shape generation is trained in three stages. First, we train a PointNet-based autoencoder for each object category using a Chamfer-distance reconstruction loss. The autoencoder uses a $512$-dimensional latent space and is trained with Adam at a learning rate of $5\times10^{-4}$ and batch size $32$. We train for $1000$ epochs on Chair and $2000$ epochs on Car and Airplane.

In the second stage, we train a context generator to map a $512$-dimensional Gaussian input to the latent space of the frozen autoencoder. The context generator uses the residual MLP described in Section~\ref{sec:appendix-arch}, with hidden dimension $1024$ and $4$ residual blocks. It is trained with MF/iMF using Adam, a learning rate of $10^{-4}$, and per-GPU batch size $16$, for up to $2000$ epochs.

In the final training stage, the point-cloud generator is conditioned on latents produced by the auto-encoder and maps source point samples to the target shape. It uses the same pointwise residual-MLP design with hidden dimension $2048$ and $12$ residual blocks. We train this model with Adam at a learning rate of $2\times10^{-5}$ and per-GPU batch size $16$, again for up to $2000$ epochs.

We use the same architecture and training configuration across all three ShapeNet categories without extensive category-specific hyperparameter tuning. The pointwise MLP provides a simple and efficient backbone that supports generation with a variable number of points, while allowing us to focus the evaluation on the proposed OT-enhanced training framework rather than on a specialized point-cloud architecture.

\subsection{Shape Completion}

For shape completion, the same model (hidden width $2048$, depth $12$) is used, with its PVCNN2-based partial-shape encoder (Section~\ref{sec:appendix-arch}) mapping a $200$-point partial/single-view input to a $256$-dimensional conditioning latent, trained jointly with the generator end-to-end. All models are trained with Adam at a learning rate of $4\times10^{-5}$, betas $(0.5, 0.999)$, and per-GPU batch size $16$, for up to $2000$ epochs, on four GPUs.

\paragraph{Shape-completion objective.}
Our completion model follows the conditional generative formulation commonly used in point-cloud completion methods such as PVD: the partial observation is provided as conditioning, while the model learns to generate points from the corresponding complete shape. Specifically, during training, the target points are sampled from the full ground-truth point cloud rather than explicitly restricted to the geometrically unobserved region. Thus, the objective learns the conditional distribution of the complete shape given the partial input, rather than requiring an explicit correspondence between generated points and a predefined missing subset.

At inference, we generate the number of points required to complete the partial input and combine them with the observed points to obtain the final completed shape. Because the generated set is not explicitly constrained to be disjoint from the observed region, some overlap with already observed geometry is possible. This is not specific to our OT-MF formulation. Established generative completion approaches similarly condition on the partial observation without explicitly enforcing a missing-region mask. The conditioning encoder instead provides the model with information about the observed geometry, while reconstruction quality is evaluated on the resulting full point cloud using standard completion metrics.

Accordingly, the term ``generating the remaining points'' in the main paper refers to generating the additional point set used to complete the partial observation, rather than explicitly generating the geometric set difference between the full and partial point clouds.

\textbf{Evaluation Protocol.} Given the $S{=}200$ observed points of a view as conditioning input, the model generates the remaining $N-S$ points of the shape ($N{=}2048$). The final reconstructed point cloud used for evaluation is the \emph{concatenation} of the real observed points with the generated points,
\[
\hat{X} = X_{\text{obs}} \cup X_{\text{gen}}, \qquad |X_{\text{obs}}| = S, \quad |X_{\text{gen}}| = N - S,
\]
so the observed geometry is preserved exactly in the reported reconstruction rather than being re-synthesized. This matches PVD's evaluation procedure, in which the observed points are held fixed and concatenated back into the point set at every step of the reverse diffusion process before scoring.
\subsection{Data Normalization}

To ensure fair comparison with prior work, we adopt the same data normalization procedure used in previous methods. We compute a global mean and standard deviation across all training samples and normalize the data during training. The same statistics are used to transform generated outputs back to the original coordinate scale. This follows the protocols used in PVD\cite{zhou20213d}, PSF\cite{10203334}, and NSOT\cite{hui2025nsot}.

\subsection{OT Solver Configuration}
\label{sec:appendix_ot_solver}

All couplings are computed on the squared Euclidean ground cost $c(x,y) = \|x-y\|_2^2$, with uniform marginals over the batch (or over the $P$ points, for the point-level coupling). All reported results use the \emph{exact} OT solver: an unregularized transport plan obtained by solving the corresponding linear program with the network-simplex algorithm (\texttt{ot.emd} from the POT library~\cite{pot}). Because the network simplex algorithm is an exact combinatorial solver rather than a fixed-point iteration, it has no precision-vs-iterations trade-off; the solver's iteration cap is set conservatively high (effectively unbounded for the problem sizes used here, $\leq 256$ points) and is never the binding constraint -- the solver always returns the LP-optimal plan.

\subsection{Computational Overhead of Point-Level OT}
\label{sec:appendix_ot_overhead}

We measure the wall-clock cost of the point-level OT solve directly inside the training loop, on our 4-GPU (DDP, one process per GPU) training configuration with a per-GPU batch of $B{=}16$ point clouds subsampled to $P{=}256$ points each. Table~\ref{tab:ot_overhead} reports the average time spent solving the $B$ independent $P\times P$ exact transport problems per training step, relative to the average total step time (forward, OT coupling, and backward pass).

\begin{table}[h]
\centering
\small
\caption{Wall-clock overhead of the point-level exact OT coupling, measured per training step on 4-GPU DDP training ($B{=}16$ point clouds/GPU, $P{=}256$ points/cloud).}
\label{tab:ot_overhead}
\begin{tabular}{@{}lccc@{}}
\toprule
Component & Time / step & \% of step & Notes \\
\midrule
Point-level OT coupling & $\sim$89\,ms & $\sim$15\% & $B$ sequential $P{\times}P$ solves \\
Total training step & $\sim$577\,ms & 100\% & fwd + OT + bwd \\
\bottomrule
\end{tabular}
\end{table}

The point-level OT coupling adds roughly 15\% to the per-step training time. This cost is dominated by the fact that the $B$ transport problems are solved sequentially on CPU (the network-simplex solver operates on NumPy arrays, requiring a device-to-host transfer of each $P{\times}P$ cost matrix), rather than being batched. For reference, the batch-level coupling used in the latent-space generator -- a single $16\times16$ problem in a 512-dimensional space -- costs under 3\,ms/step, roughly $30\times$ cheaper in absolute terms despite having a comparable relative overhead ($\sim$11\% of that stage's much shorter step time), because it requires only one exact solve per step instead of $B$.

\subsection{Discussion of Latent-Space Transport}
\label{sec:latent_ot_discussion}

Our latent-space shape transport should be viewed as a scalable
approximation to outer OT in the original point-cloud space, rather
than an equivalent reformulation. In the current implementation, the
autoencoder is trained with a Chamfer reconstruction loss, which does
not explicitly enforce that Euclidean distances between latent
representations preserve Wasserstein distances between the
corresponding point-cloud distributions. Therefore, the latent-space
OT coupling is not guaranteed to coincide with the coupling obtained
from exact outer OT using permutation-aware point-cloud costs.

A promising direction is to learn a Wasserstein-aware latent geometry.
For example, Wasserstein Wormhole~\cite{pmlr-v235-haviv24a} learns
point-cloud embeddings whose Euclidean distances approximate pairwise
Wasserstein distances. Incorporating a similar distance-preserving
objective into our encoder could make latent-space transport more
faithful to the original point-cloud OT geometry while retaining its
computational advantages. Establishing formal guarantees for this
approximation, and studying Wasserstein-geometric encoders in
OT-MF3D, are left for future work.

\section{Failure Cases}

\begin{figure}
  \centering  
  \includegraphics[width=\textwidth]{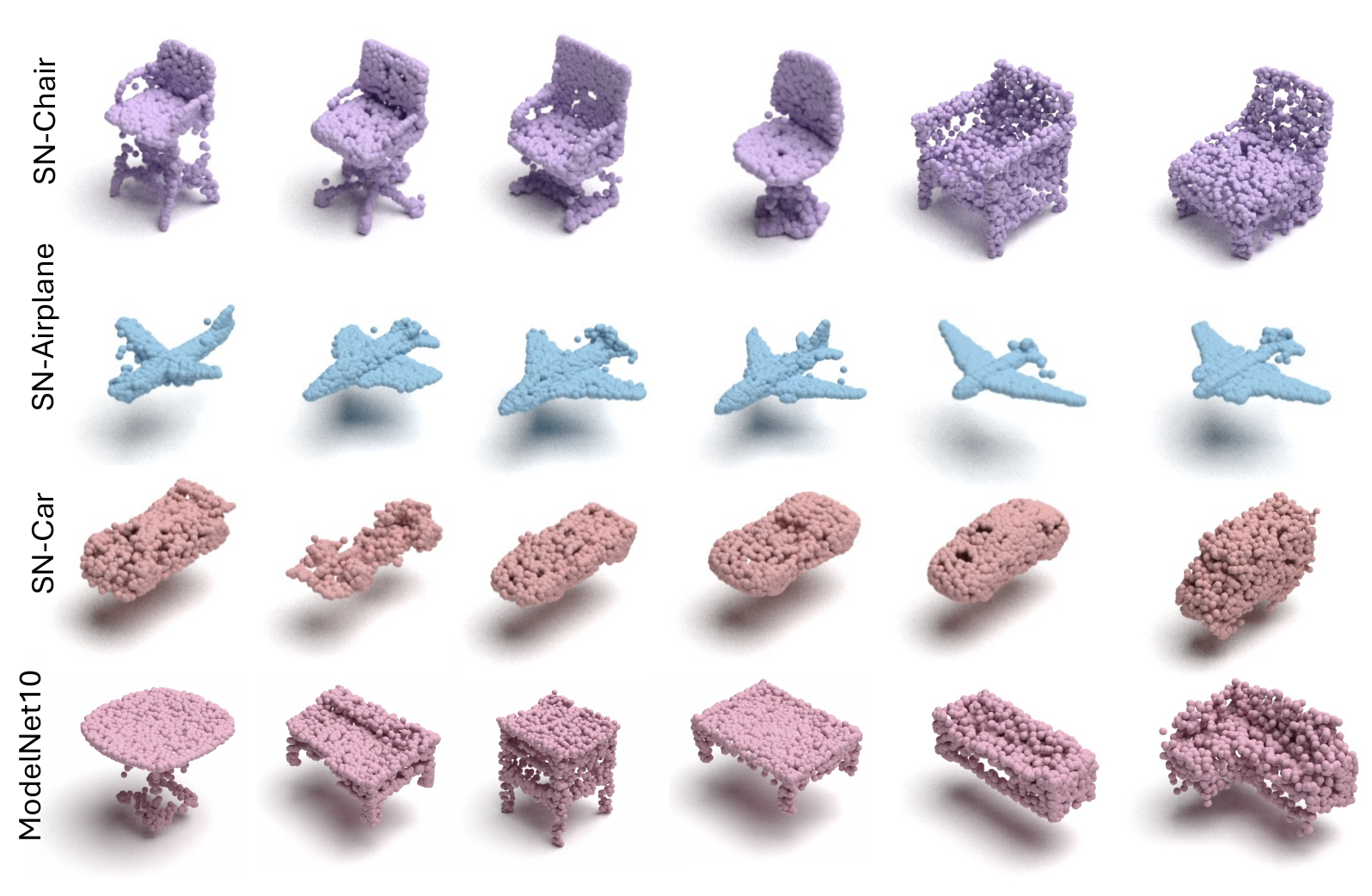}
  \caption{Some failure cases of our method across different datasets.
  }
  \label{fig:failure_cases}  
\end{figure}

Figure~\ref{fig:failure_cases} presents representative failure cases of our one-step OT-MF3D model. 
Overall, the generated samples typically preserve the coarse geometry and semantic structure of the target object category, while the remaining errors are primarily fine-scale details.

For SN-Chair and ModelNet10, the model can struggle with thin or fine-grained structures. 
For example, chair legs, backrest, and other small components may become incomplete, or disconnected. 
Similar behavior is observed for some ModelNet10 objects, where the overall object shape remains recognizable but local support structures and geometric details can be poorly resolved.
In contrast, SN-Airplane and SN-Car generally exhibit more consistent global shapes. 
Their failures are more commonly shown by small amounts of noise, such as detached points around airplane wings or tails and local irregularities around car surfaces. 

One-step generation is a particularly challenging setting since the complete transformation from the initial distribution to the data distribution must be performed in a single generation step.
Moreover, while the point-level OT matching provides strong geometric correspondences during training, it does not explicitly enforce properties such as part connectivity or local surface smoothness.
Therefore, small structural inconsistencies can remain in the generated point clouds.

\section{Decoder comparison}

To compare the OT-MF based decoder with the original decoder from the initial pre-trained AE, we compare their performance under a unified setting. Both models are conditioned with the latent vector generated from the same context generator.
Tables~\ref{tab:decoder_otmf_1nn} and~\ref{tab:decoder_otmf_mmd_cov} compare samples generated by OT-MF against those obtained directly from the pretrained decoder. Overall, OT-MF provides a substantially better match to the target shape distribution, particularly for the EMD-based metrics. For 1-NN-EMD, OT-MF consistently moves the accuracy much closer to the ideal value of $50\%$ across all three categories, reducing the average from $85.29$ for the decoder to $57.38$. A similar trend is observed for coverage: OT-MF improves EMD-COV from $26.42$ to $48.15$ on Airplane, from $33.53$ to $52.11$ on Chair, and from $28.69$ to $50.57$ on Car. OT-MF also achieves lower EMD-MMD across all categories. These results show the necessity to train our OT-MF shape generator. Although the decoder can generate samples with competitive nearest-neighbor distances under CD, OT-MF captures the overall data distribution much better.

\begin{table*}[t]
\centering
\caption{Shape generation results on ShapeNet-Airplane, Chair, and Car using
1-NN accuracy. Values closer to $50$ are better. Best results between the
Decoder and OT-MF are shown in \textbf{bold}.}
\label{tab:decoder_otmf_1nn}

\setlength{\tabcolsep}{8pt}
\begin{tabular}{l cc cc cc cc}
\toprule
\multirow{2}{*}{Model}
& \multicolumn{2}{c}{Airplane}
& \multicolumn{2}{c}{Chair}
& \multicolumn{2}{c}{Car}
& \multicolumn{2}{c}{Avg.} \\

\cmidrule(lr){2-3}
\cmidrule(lr){4-5}
\cmidrule(lr){6-7}
\cmidrule(lr){8-9}

& CD & EMD
& CD & EMD
& CD & EMD
& CD & EMD \\

\midrule

Decoder
& 71.73 & 93.70
& 57.48 & 77.49
& \textbf{57.10} & 84.66
& \textbf{62.10} & 85.29 \\

OT-MF
& \textbf{71.60} & \textbf{64.07}
& \textbf{57.40} & \textbf{54.08}
& 58.81 & \textbf{53.98}
& 62.60 & \textbf{57.38} \\

\bottomrule
\end{tabular}
\end{table*}

\begin{table*}[t]
\centering
\scriptsize
\caption{Shape generation results on ShapeNet-Airplane, Chair, and Car with
MMD and COV scores. CD-MMD is multiplied by $10^{3}$ and EMD-MMD is
multiplied by $10^{2}$, while COV is reported as a percentage.
Best results between the Decoder and OT-MF are shown in \textbf{bold}.}
\label{tab:decoder_otmf_mmd_cov}

\resizebox{\textwidth}{!}{
\begin{tabular}{l cccc cccc cccc}
\toprule

\multirow{3}{*}{Model}
& \multicolumn{4}{c}{Airplane}
& \multicolumn{4}{c}{Chair}
& \multicolumn{4}{c}{Car} \\

\cmidrule(lr){2-5}
\cmidrule(lr){6-9}
\cmidrule(lr){10-13}

& \multicolumn{2}{c}{MMD$\downarrow$}
& \multicolumn{2}{c}{COV$\uparrow$}
& \multicolumn{2}{c}{MMD$\downarrow$}
& \multicolumn{2}{c}{COV$\uparrow$}
& \multicolumn{2}{c}{MMD$\downarrow$}
& \multicolumn{2}{c}{COV$\uparrow$} \\

\cmidrule(lr){2-3}
\cmidrule(lr){4-5}
\cmidrule(lr){6-7}
\cmidrule(lr){8-9}
\cmidrule(lr){10-11}
\cmidrule(lr){12-13}

& CD & EMD & CD & EMD
& CD & EMD & CD & EMD
& CD & EMD & CD & EMD \\

\midrule

Decoder
& \textbf{0.195} & 0.576
& 47.65 & 26.42
& \textbf{2.360} & 1.743
& 45.02 & 33.53
& \textbf{0.814} & 0.933
& 42.33 & 28.69 \\

OT-MF
& 0.229 & \textbf{0.370}
& \textbf{48.40} & \textbf{48.15}
& 2.539 & \textbf{1.487}
& \textbf{45.62} & \textbf{52.11}
& 0.910 & \textbf{0.736}
& \textbf{46.02} & \textbf{50.57} \\

\bottomrule
\end{tabular}
}
\end{table*}

\section{Additional Shape Generation Results}

Following recent work\cite{10203334, zeng2022lion}, we report the Maximum Mean Discrepancy (MMD) and coverage (COV) metrics for evaluating generative performance. Table \ref{tab:shapenet_mmd_cov} presents the results for shape generation on the ShapeNet datasets. Our method achieves the best or second-best performance on the majority of datasets across both metrics, specifically compared with one-step generation approaches such as PSF\cite{10203334} and 1-GAN\cite{pmlr-v80-achlioptas18a}.

Furthermore, our method achieves competitive performance relative to LION\cite{zeng2022lion}, which requires $1000$ sampling steps, while maintaining significantly more efficient training and sampling. In particular, LION requires approximately $550$ GPU hours for training\cite{zeng2022lion}, whereas our method remains substantially more efficient.

\begin{figure}
  \centering  
  \includegraphics[width=\textwidth]{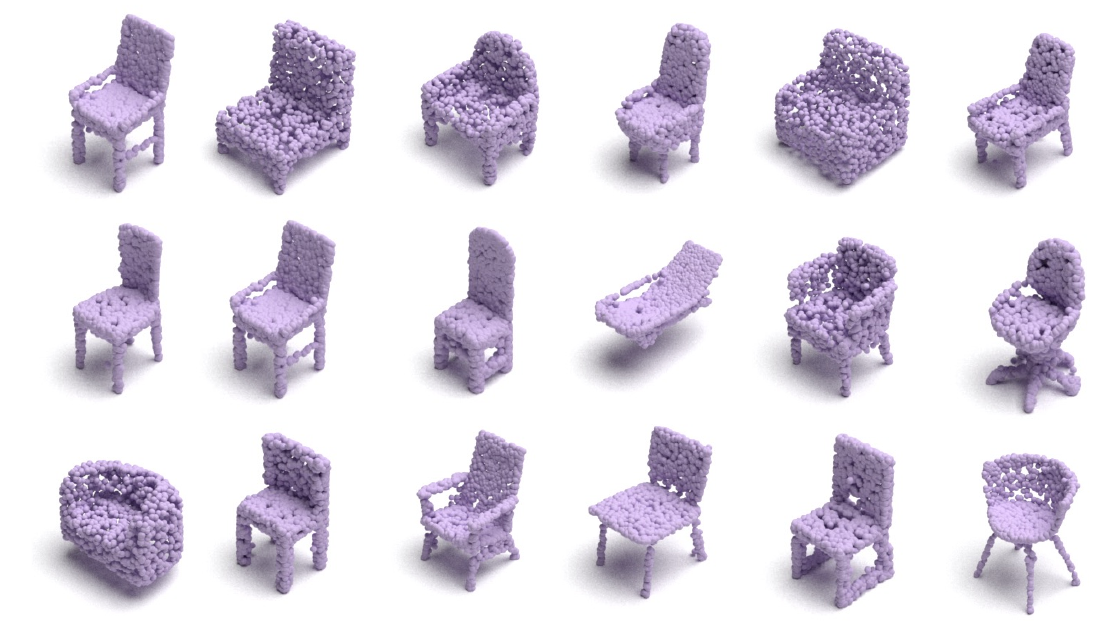}
  \caption{Additional one-step generation samples on class Chair using our model.
  }
  \label{fig:chair_extra}  
\end{figure}

\begin{figure}
  \centering  
  \includegraphics[width=\textwidth]{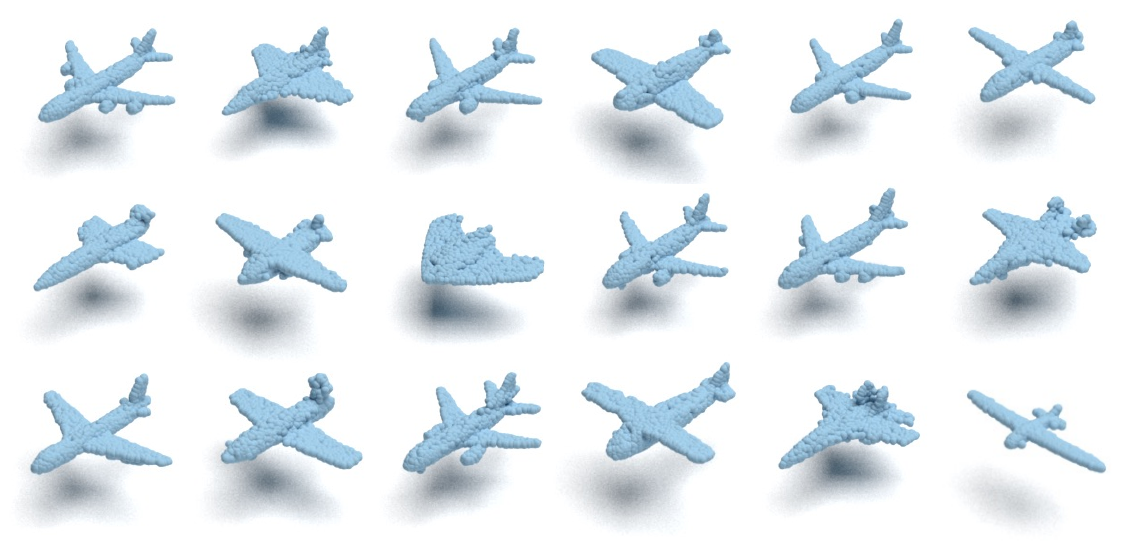}
  \caption{Additional one-step generation samples on class Airplane using our model.
  }
  \label{fig:airplane_extra}  
\end{figure}

\begin{figure}
  \centering  
  \includegraphics[width=\textwidth]{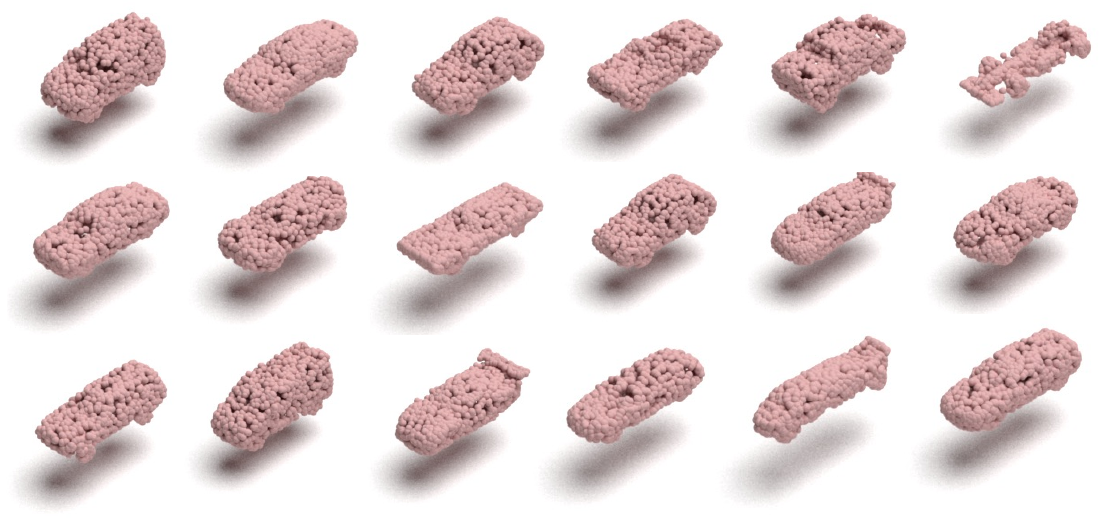}
  \caption{Additional one-step generation samples on class Car using our model.
  }
  \label{fig:car_extra}  
\end{figure}

In addition, Figures~\ref{fig:chair_extra}, \ref{fig:airplane_extra}, and \ref{fig:car_extra} present additional one-step generation samples produced by our method for the Chair, Airplane, and Car datasets, respectively. As illustrated in these figures, our approach consistently generates diverse and high-quality shapes in a single step.

\begin{table}
\centering
\scriptsize
\caption{Shape generation results on ShapeNet-Airplane, Chair, and Car with MMD and COV scores. CD is multiplied by $10^{3}$ and EMD is multiplied by $10^{2}$.}

\resizebox{\textwidth}{!}{
\begin{tabular}{l|cccc|cccc|cccc}
\toprule
\multirow{3}{*}{Model} 
& \multicolumn{4}{c|}{Airplane} 
& \multicolumn{4}{c|}{Chair} 
& \multicolumn{4}{c}{Car} \\

\cmidrule(lr){2-5}
\cmidrule(lr){6-9}
\cmidrule(lr){10-13}

& \multicolumn{2}{c}{MMD$\downarrow$} 
& \multicolumn{2}{c|}{COV$\uparrow$}
& \multicolumn{2}{c}{MMD$\downarrow$} 
& \multicolumn{2}{c|}{COV$\uparrow$}
& \multicolumn{2}{c}{MMD$\downarrow$} 
& \multicolumn{2}{c}{COV$\uparrow$} \\

\cmidrule(lr){2-3}
\cmidrule(lr){4-5}
\cmidrule(lr){6-7}
\cmidrule(lr){8-9}
\cmidrule(lr){10-11}
\cmidrule(lr){12-13}

& CD & EMD & CD & EMD
& CD & EMD & CD & EMD
& CD & EMD & CD & EMD \\

\midrule

1-GAN (CD) \cite{pmlr-v80-achlioptas18a} 
& 0.340 & 0.583 & 38.52 & 21.23 
& 2.589 & 2.007 & 41.99 & 29.31 
& 1.532 & 1.226 & 38.92 & 23.58 \\

1-GAN (EMD) \cite{pmlr-v80-achlioptas18a}  
& 0.397 & 0.417 & 38.27 & 38.52 
& 2.811 & 1.619 & 38.07 & 44.86 
& 1.408 & 0.899 & 37.78 & 45.17 \\

PointFlow \cite{yang2019pointflow}
& 0.224 & 0.390 & 47.90 & 46.41
& \textbf{2.409} & 1.595 & 42.90 & 50.00
& \textbf{0.901} & 0.807 & 46.88 & 50.00 \\

SoftFlow \cite{10.5555/3495724.3497099}
& 0.231 & 0.375 & 46.91 & 47.90
& \underline{2.528} & 1.682 & 41.39 & 47.43
& 1.187 & 0.859 & 42.90 & 44.60 \\

DPF-Net \cite{10.1007/978-3-030-58592-1_41}
& 0.264 & 0.408 & 46.17 & 48.89
& 2.536 & 1.632 & 44.71 & 48.79
& 1.129 & 0.853 & 45.74 & 49.43 \\

Shape-GF \cite{10.1007/978-3-030-58580-8_22}
& 2.703 & 0.659 & 40.74 & 40.49
& 2.889 & 1.702 & 46.67 & 48.03
& 9.232 & 0.756 & \underline{49.43} & 50.28 \\

PVD \cite{zhou20213d}
& 0.224 & 0.380 & \textbf{48.88} & \underline{52.09}
& 2.622 & \underline{1.556} & \underline{49.84} & 50.60
& 1.077 & 0.794 & 41.19 & 50.56 \\

PVD-DDIM (N=100) \cite{10203334}
& 0.243 & 0.399 & 44.23 & 49.75
& 2.758 & 1.703 & 46.32 & 48.19
& 1.202 & 0.818 & 40.01 & 48.34 \\

PSF (N=1) \cite{10203334}
& \underline{0.221} & \underline{0.366} & 46.17 & \textbf{52.59}
& 2.624 & 1.573 & 46.71 & 49.84
& 1.023 & 0.802 & 42.89 & \underline{53.12} \\

LION (N=1000) \cite{zeng2022lion}
& \textbf{0.219} & 0.372 & 47.16 & 49.63
& 2.640 & 1.550 & 48.94 & \textbf{52.11}
& 0.913 & 0.752 & \textbf{50.00} & \textbf{56.53} \\

\midrule
\textbf{OT-MF3D (N=1)}
& \underline{0.221} & \textbf{0.360} & 47.90 & 48.39
& 2.656 & \textbf{1.504} & \textbf{50.75} & \underline{51.81}
& 0.918 & \underline{0.747} & 46.59 & 48.86 \\
\textbf{OT-MF3D(imp)(N=1)}
& 0.229 & 0.370 & \underline{48.39} & 48.14
& 2.538 & 1.486 & 45.61 & 52.11
& \underline{0.910} & \textbf{0.736} & 46.02 & 50.57 \\

\bottomrule
\end{tabular}
}
\label{tab:shapenet_mmd_cov}
\end{table}

\end{document}